%% file: main.tex
\RequirePackage{fix-cm}
\documentclass{svjour3}                     % onecolumn (standard format)
\smartqed  % flush right qed marks, e.g. at end of proof
\usepackage{graphicx}
\usepackage{graphicx}
\usepackage{amssymb}
\usepackage{amsmath}
\usepackage{bbding}
\usepackage{url}
\usepackage[ruled,lined,titlenumbered,linesnumbered]{algorithm2e}
\usepackage{hyperref}
\usepackage{stmaryrd}
\usepackage{rotating}
\usepackage[sectionbib,round]{natbib}

\journalname{Journal of Intelligent Manufacturing}
\begin{document}

\title{Metaheuristics for the Template Design Problem: Encoding, Symmetry and Hybridisation%\thanks{Grants or other notes
}

\titlerunning{Metaheuristics for the Template Design Problem}        % if too long for running head

\author{David Rodr\'{i}guez Rueda \and
	Carlos Cotta \and
	Antonio J. Fern\'{a}ndez-Leiva$^*$
}


\institute{D. Rodriguez \at
	Universidad Nacional Experimental del T\'{a}chira (UNET), Laboratorio de Computaci\'{o}n de Alto Rendimiento (LCAR), San Crist\'{o}bal, T\'{a}chira, 5001, Venezuela \\
	\email{drodri@unet.edu.ve}           %  \\
	\and
	C. Cotta \at
	ITIS Software, Universidad de M\'{a}laga, ETSI Inform\'{a}tica, Campus de Teatinos, 29071 M\'{a}laga, Spain \\
	ORCID number:  0000-0001-8478-7549\\
	\email{ccottap@lcc.uma.es}. 
	\and
	$^*$ A.J. Fern\'{a}ndez-Leiva (Corresponding author) \at
	ITIS Software, Universidad de M\'{a}laga, ETSI Inform\'{a}tica, Campus de Teatinos, 29071 M\'{a}laga, Spain \\
	ORCID number: 0000-0002-5330-5217\\
	\email{afdez@lcc.uma.es} \\
	Tel.: $+34\ 952133315$
}

\date{Received: date / Accepted: date}

\maketitle

\begin{abstract}
The template design problem (TDP) is a hard combinatorial problem with a high number of symmetries which makes  solving it  more complicated. A number of techniques have been proposed in the literature to optimise its resolution, ranging from complete methods to stochastic ones. However, although metaheuristics are considered efficient methods that can find enough-quality solutions at a reasonable computational cost, these techniques have not proven to be truly efficient enough to deal with this problem. This paper explores and analyses a wide range of metaheuristics to tackle the problem with the aim of assessing  their suitability for finding template designs. 

We tackle the problem using a wide set of metaheuristics whose implementation is  guided by a number of issues such as  problem formulation, solution encoding, the symmetrical nature of the problem, and distinct forms of hybridisation.
 For the TDP, we also propose a slot-based alternative problem formulation (distinct to other slot-based proposals), which represents  another option other than the classical variation-based formulation of the problem.

An empirical analysis, assessing the performance of all the metaheuristics  (i.e.,  basic, integrative and collaborative algorithms  working on different search spaces and with/without symmetry breaking) shows that some of our proposals can be considered the state-of-the-art when they are applied to specific problem instances. 
\end{abstract}

\input{introduction}

\input{background}

\input{otherPerspectivesTotackleTDP}

\input{hybridalgorithms}

\input{experiments}

\input{conclusions}

\begin{acknowledgements}
This work is partially funded by Ministerio Espa\~{n}ol de Econom\'{\i}a y Competitividad (projects TIN2014-56494-C4-1-P, UMA::EPHEMECH -- \url{https://ephemech.wordpress.com/} and TIN2017-85727-C4-1-P, UMA::DeepBio -- \url{https://deepbio.wordpress.com/}) and Universidad de M\'{a}laga.
\end{acknowledgements}

\newcommand{\noopsort}[1]{} \newcommand{\printfirst}[2]{#1}
  \newcommand{\singleletter}[1]{#1} \newcommand{\switchargs}[2]{#2#1}

\end{document}

%% file: introduction.tex
\section{Introduction}
\label{Introduction}
Many problems in the area of manufacturing are related to reducing the waste of the raw material used in the production process \citep{Kasemset20151342,wang2016waste}. In general, achieving this objective requires a huge effort in the analysis of the problem in order to obtain a suitable model that allows greater production with minimum waste. The template design problem (TDP) is a challenging example of this. The TDP arises in industrial settings in which variations of a given product must be produced, each of them requiring a particular packaging (typically with different printing patterns). The production of these packages entails minimising the use of cardboard (or any other raw material used). Appropriate templates for printing these packages must therefore be designed, hence the TDP.

The TDP was first described by Proll and Smith (\citeyear{Proll97ilpand}) who observed this problem arising at a local colour printing firm. Roughly speaking, we can assume a certain product has to be manufactured with distinct variations (e.g., different flavours of cereal flakes), each one requiring a similar --but different-- packaging. A printing machine is used to produce this packaging. This machine is configured with a given template, which is subsequently pressed on sheets of raw material (e.g., cardboard). Given the large number of items required, a template comprises several slots, each of them filled with a given variation of the product, which are printed on each pressing. In addition, there can be more than one such template. This means that the problem is twofold: (i) determine the design of each template, namely which variations are included in each slot, and (ii) determine the optimal usage of these templates. The latter requires a given criterion to be optimised, for example minimising the manufacturing time (i.e., minimising the number of pressings) or minimising the waste, given the known demands of each variation. We consider here the latter criterion (i.e., optimise the use of raw material). 

In the literature, one can find various  proposals that deal with the TDP, including constraint programming techniques, mathematical programming and integer linear programming. 
 Proll and Smith used an integer linear model to solve this problem. It must be noted that the problem is 
intrinsically symmetrical in nature, meaning that one solution can be represented in different ways. This can exert
an influence on the way the search is conducted (and ultimately on the performance of the algorithm). Indeed, it has been shown that an adequate treatment of such symmetries with symmetry-breaking techniques can reduce the complexity of the search \citep{Janyen2016}. In this sense, one can consider the equivalence among solutions from different perspectives. For instance, with a numeric value \citep{benhamou1994study} or with a geometric approach  \citep{backofen2002excluding}, just to name a couple
 In the last few decades, a number of methods have been applied to deal with this interesting issue \citep{benhamou1994study,fahle2001,Gent99symmetrybreaking}. The primary method, in constraint and integer programming, to cope with symmetries consists in breaking them, i.e., removing symmetries with the goal of reducing the search space of the problem. 

With regard to  metaheuristics, there are not many proposals that deal with the TDP, although it has been proved that metaheuristics are efficient methods in the solving of manufacturing problems (e.g., \citep{DBLP:journals/jim/MeeranM12,DBLP:journals/jim/LinC18}).
More recently, we handled the problem with some basic metaheuristics (i.e., local searches, and genetic algorithms)  \citep{RodriguezCottaFernandezMAEB2010, RodriguezCottaFernandezIMACS2011}. These techniques demonstrated a moderate success as they performed reasonably well for small instances of the problem but their performance worsened when they were applied to more complex instances. In fact, the problem formulation is ideal for the employment of integer linear programming (ILP) techniques, as it has been proven in the literature. However, metaheuristics can also be suitable for tackling the problem as they offer a worthwhile balance between the quality of the solutions found and the computational cost to find them. 
The main contribution of this paper is to explore this issue, that is to say, the goodness of metaheuristics to tackle the TDP. Thus, this paper contains the description of a wide range of  metaheuristics, of distinct nature,  to cope with the problem. Our proposals can also be viewed as an  alternative mechanism to those already reported in the literature, to tackle the TDP. 

Thus, in general, this paper tries to shed light on the solving of the TDP using metaheuristics, and considers three main issues for the design of these techniques: problem  symmetry, problem formulation (and related aspects such as the search space representation), and hybrid forms of collaboration between metaheuristics (i.e., integrative vs. cooperative schemes).
 So, this paper firstly considers a standard procedure, used in constraint and integer programming,  for symmetry breaking, that is to say,  the addition of new constraints to the problem with the aim of removing symmetries and ease its solving. 
 Note however, that other mechanisms of dealing with symmetries have been proposed for genetic algorithms and local search; for instance, in \cite{DBLP:journals/tec/Prugel-Bennett04}
symmetry breaking is modelled using stochastic differential equations and their associated diffusion equations.
Secondly,  based on the assumption that the candidate encoding  can drastically affect the search process, this paper also considers an alternative integer slot-based  representation scheme for the TDP solutions (that we have called the \emph{alternative} model, in response to the variation-based model that has been classically taken  as reference). Note however that our proposal is not the first slot-based scheme proposed to deal with the problem; a 0/1 variable slot-based approach was employed in \citet{prestwich06}. 
In this context, we describe a number of optimisation methods to deal with TDP that are derived from  all the possible scenarios that  arise from the combination of these two encodings  and the decision on whether to apply symmetry breaking. Each scenario is firstly tackled with a number of basic metaheuristics, including local search (LS) and genetic algorithms (GA).

In addition, it has been proven that the use of hybrid algorithms represents a very strong mechanism for improving the search capability of optimisation algorithms \citep{Ting2015}. Generally speaking, hybridisation can be viewed from two broad point of view \citep{Raidl2006}:  integration and cooperation. Integration usually refers to the adding of one optimisation technique as a component of another optimisation method, whereas cooperation is generally related to the establishment of a way to exchange information between methods that are applied one after another or in parallel. Meanwhile, \citet{Crainic2003}, consider hybridisation basically as a synergistic union of different algorithmic approaches, such that at least one of them represents an exploitation mechanism of knowledge. In this paper, we also explore this path, and use an integrative mechanism which embeds a  local search (LS) inside a genetic algorithm (GA) resulting in a memetic approach (MA) \citep{Neri2012book}. We also develop cooperative algorithms in which both the basic and integrative metaheuristics (i.e., LSs, GAs and MAs) work independently to handle the problem and interchange information in certain synchronisation moments that have been previously preset. All the hybrid metaheuristics described in this paper are applied to solving the TDP for first time (to the best of our knowledge). One of the novelties of these hybrid methods  is that they are allowed to link metaheustics that are very different  to each other in the sense that the connected methods can vary in their encoding schemes, the  problem formulation that is handled, the use or absence  of constraints for symmetry breaking, and/or even the nature of the method (e.g., an LS or a MA).  We have also conducted an experimental evaluation and have compared the performance of all the metaheuristics proposed here. The results show that some of our metaheuristics optimisation methods for the TDP can be considered the  state-of-the-art when they are applied to specific problem instances of the problem. Thus, the main contribution of this paper is to show the goodness of metaheuristics to address the TDP. In addition, the paper suggests that other possible forms of hybrid metaheuristics could be suitable for the solving and optimisation of TDPs, problems that  have traditionally been efficiently tackled by ILP methods.

%% file: background.tex
\section{Background}

This section first describes the classical formulation of the problem. Then, we discuss a number of approaches tackling the TDP that have been reported in the literature.

\input{formulation}

\input{related}

%% file: formulation.tex
\subsection{TDP: Formulation and classical model}
\label{subsec:primalmodel}
\sloppypar As mentioned in the introduction,  the TDP was first described by Proll and Smith (\citeyear{Proll97ilpand}) who observed this problem arising at a local colour printing firm. They used an integer linear programming (ILP) formulation 
to  determine the optimal usage of the available templates. Let $V$ be a set of $v$ variations to
be produced and  $T$ a set of $t$ templates $T_1,\cdots,T_t$ (each of them with $s$ slots). Note that the particular slot in which a
variation is placed is irrelevant: it only matters how many
instances of a certain variation are contained in a given template.
Thus, let $s_{ij}$ be the number of instances of variation $i$ in
template $T_j$ (in other words, the number of slots in template $j$ in which variation $i$ appears). Now, let $Q_i$ be the demand for variation $i$ (deterministic
and known; see \citep{prestwich06} for an approach under uncertainty), and
let us assume that we have production tolerances $l_i,u_i \in [0.0,1.0]$ for each
variation $i$, i.e. we can permit up to a certain underproduction
$Q_i(1-l_i)$ or overproduction $Q_i(1+u_i)$ for each variation $i$. Then,
the resulting problem is formulated as follows:

\begin{equation}
\varphi = \min{\sum_{i=1}^v\left(U_i+O_i\right)}
\end{equation}
subject to:
\begin{equation} %
\label{primal model: constraint 1}
\sum_{j=1\ldots t}s_{ij}R_j+U_i-O_i = Q_i,\ \ 1 \leqslant i
\leqslant v
\end{equation}
\begin{equation}
\label{primal model: constraint 2}
\sum_{j=1\ldots t}s_{ij}R_j \geqslant (1-l_i)Q_i,\ \ 1 \leqslant i
\leqslant v
\end{equation}
\begin{equation}
\label{primal model: constraint 3}
\sum_{j=1\ldots t}s_{ij}R_j \leqslant (1+u_i)Q_i,\ \ 1 \leqslant i
\leqslant v
\end{equation}
\begin{equation}
R_j \geqslant 0,\ \ 1 \leqslant j \leqslant t
\end{equation}
\begin{equation}
U_i,O_i \geqslant 0,\ \ 1 \leqslant i \leqslant v
\end{equation}
where $U_i$ and $O_i$ are slack variables that respectively represent the underproduction and overproduction of variation $i$, and $R_j$ denotes the number of times template $j$ is punched (pressed). Note that  $\varphi $ is a feasible solution if: 

$$(1-l_i)Q_i \le \varphi \le (1+u_i)Q_i$$

Originally, Proll and Smith considered a  lower tolerance limit $f_\%$ of 10\%, so (in the experiments described in Section \ref{sec:experiments}) we maintain this value for the under and the over production (i.e., $l_i=u_i=0.10$ so that  $f_\% = 10\%$).

In the rest of the paper,  $\text{TDP}\langle v,t,s\rangle$ denotes a TDP instance  with $v$ variations, $t$ templates and $s$ slots per template.

%% file: related.tex
\subsection{Related work}
\label{sect:relatedwork}
The TDP problem has been tackled with a number of different techniques in the literature, with differing levels of success. Traditionally,
the problem was treated via deterministic, constructive and/or complete methods. Proll and Smith (\citeyear{Proll97ilpand}) addressed this problem for the first time using the formulation  shown in Section \ref{subsec:primalmodel}. Their proposal was based on integer linear programming and constraint programming, and very promising results were obtained. They also concluded that addressing this problem represents a  very difficult task. Subsequently, \citet{Pierre01} presented a matrix model to tackle the problem. This approach suggests first fixing  the number of templates to later minimise the total number of pressings. They employed a two-dimensional array to store the number of copies of each variation that were being used in each template. Later, \citet{prestwich06} described an ILP-based local search algorithm (called VWILP), which was based on a mechanism of uncertain demand. The algorithm was also based on a state-of-the-art SAT (Boolean satisfiability) method.

With regards to metaheuristics, \citet{RodriguezCottaFernandezMAEB2010} proposed the use of two local search (LS)  algorithms, namely, a simple scheme based on local search for maximum slope (hill climbing, HC), together with a more robust technique based on tabu search (TS). Here, taking the model suggested in \cite{Pierre01}, each candidate solution was represented by a matrix $M=\{s_{ij}\}_{v\times t}$ where each $s_{ij} \in \{1..s\}$ stores the number of slots in template {$j$} in which variation {$i$} appears. 
Later, in \citep{RodriguezCottaFernandezIMACS2011}, a genetic algorithm (GA) for handling the problem was described. This GA used binary tournament for selection, a $(\mu+1)$--policy for replacement (i.e., a new individual is generated and inserted in the population replacing the worst one) and breeding was done by recombination and mutation as usual. The recombination operator was a variant of uniform crossover (UX)  \citep{Syswerda89} where a template-level exchange was carried out. 
Additionally, a greedy recombination operator (Gd) was considered. The mutation was handled in the same way as in neighbourhood  in \citep{RodriguezCottaFernandezMAEB2010} (i.e., add a slot to a design in a template and decrease by one unit the number of slots associated with any other design). Other relevant parameters for GA were \emph{population size}$=100$, crossover and mutation probabilities $p_X=.9$ and $p_M=1/(vt)$.

An experimental analysis on several instances of the problem showed that  the population-based approach performed better than the LS approaches previously considered.

%% file: otherPerspectivesTotackleTDP.tex
\section{Other perspectives to address the problem}
\label{Other perspectives}
In this section we propose to tackle the problem from distinct perspectives to those already reported in the literature. 
So, in Section~\ref{sec:DualModel} an alternative slot-based problem representation, which is different to the slot-based representation proposed in \cite{prestwich06},
 and a new problem formulation based on it, are described. Then, Section~\ref{sec:symetriesbreaking} proposes applying a standard symmetry-breaking method 
for both the classical representation (i.e., based on variations as described in Section \ref{subsec:primalmodel}) and its  alternative encoding as described in Section~\ref{sec:DualModel}. 

From these proposals, we subsequently describe a number of new metaheuristics  for handling the TDP, which are detailed later in this paper and evaluated in the experiments described in Section~\ref{sec:experiments}.

\input{alternative}

\input{symmetrybreaking}

%% file: alternative.tex
\subsection{An alternative representation}
\label{sec:DualModel}
\begin{figure}[htb]
\begin{center}
\includegraphics[width=0.3\textwidth]{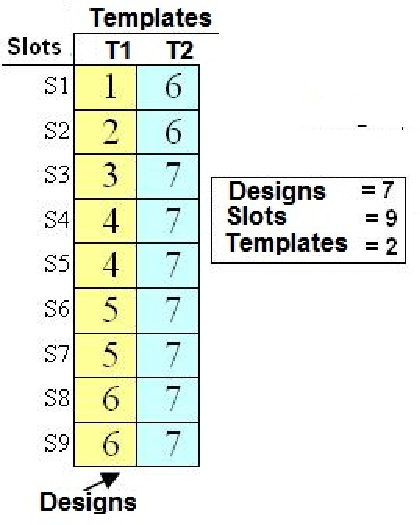}
\end{center}
\caption{Alternative model: Example of candidate encoding for the problem  instance $\text{TDP}\langle 7,2,9\rangle$.}
\label{fig:TDPModB}

\end{figure}

It is well known that the representation of candidate solutions can have a dramatic effect on the resolution of the problem, particularly in evolutionary algorithms~(\citealt{DBLP:books/daglib/0014740}; \citealt{Rothlauf:2017:REA:3067695.3067718}). Thus, here an alternative to the primary model described in Section \ref{subsec:primalmodel} is proposed. This new model is termed {\em the alternative formulation model}, denoted with the letter {\em D} because in the encoding of candidate solutions we record designs. So, in this model,  an eventual candidate is an array %$M_{SxT}$  in which 
$M_D=\{v_{ij}\}_{s\times t}$ where each $v_{ij}$ 
contains the design (i.e., variation) that is  placed in slot $i$ in a given template $j$.   
Figure~\ref{fig:TDPModB} displays an example of a possible candidate encoding for a problem instance $\text{TDP}\langle 7,2,9\rangle$.

This new slot-based formulation requires replacing the constraints shown in Equations (\ref{primal model: constraint 1})-(\ref{primal model: constraint 3}) with the following:

\begin{equation} %
\sum_{j=1\ldots t} \big(\sum_{h=1\ldots s}\llbracket v_{hj}=i\rrbracket\big)R_j+U_i-O_i = Q_i,\ \ 1 \leqslant i
\leqslant v
\end{equation}
\begin{equation}
\sum_{j=1\ldots t}(\sum_{h=1\ldots s}\llbracket v_{hj}=i\rrbracket) R_j \geqslant (1-l_i)Q_i,\ \ 1 \leqslant i
\leqslant v
\end{equation}
\begin{equation}
\sum_{j=1\ldots t}(\sum_{h=1\ldots s}\llbracket v_{hj}=i\rrbracket)R_j \leqslant (1+u_i)Q_i,\ \ 1 \leqslant i
\leqslant v
\end{equation}

\noindent where $\llbracket \cdot \rrbracket$ is the Iverson bracket, namely [\texttt{true}]$=1$ and [\texttt{false}]$=0$. Note that, for a given variation $i$ and template $j$, the expression $(\sum_{h=1\ldots s}\llbracket v_{hj}=i\rrbracket)$ returns the number of times that variation $i$ appears in template $j$.

Note that \citet{prestwich06} also employed a slot-based approach, although it used 0/1 variables instead of integers, as suggested in this paper. 

%% file: symmetrybreaking.tex
\subsection{Symmetry breaking}
\label{sec:symetriesbreaking}

Symmetry breaking is a standard technique that is well known in the scientific community. According to  Fahle, Schamberger and Sellman (\citeyear{fahle2001}), one way to reduce symmetries is to turn the problem to be solved into another problem with the same characteristics of the original but eliminating all or most symmetrical states.  In the last few decades, a number of methods have been applied to address this interesting issue \citep{benhamou1994study,fahle2001,Gent99symmetrybreaking}. The primary method, in constraint and integer programming, to cope with symmetries consists in applying symmetry-breaking techniques, that is to say,  removing symmetries with the goal of reducing the search space of the problem. Symmetry breaking can be applied in many different ways \citep{Ward-Cherrier2016,7176793,Gigliotti2015995}. 
Other recent  approaches have shown how solving combinatorial problems with mixed integer linear programming approaches can be sped up by adding symmetry-breaking constraints to the original formulation. 

Another approach, related to problem representation (one of our aforementioned issues) and, to a certain extent, to symmetry breaking, is to consider asymmetric representative formulations (ARFs) as alternatives to the natural symmetric formulation of the problem. From a general perspective, ARFs refer to choosing alternative variables to represent a problem.  ARFs have been shown to be effective in dealing with distinct combinatorial optimisation problems such as problems of job grouping, binary clustering, node coloring, or experimental designs  (\citealt{CAMPELO20081097}; \citealt{G201044}; \citealt{JANS20131132}; \citealt{G2016117}). The search of ARFs for the TDP could be an interesting line for future work. 

In this paper, and in order to reduce the search space, we have considered  a standard symmetry-breaking procedure consisting in the addition of new constraints to the problem with the aim of removing symmetries and speeding up the search process.
 The mechanisms proposed depend on the representation of the solutions/candidates, as described below for each of the two encodings considered in this paper. 
Note however that symmetry breaking is a mechanism that has already been applied to the template design \citep{Pierre01} so our proposals are specific to our problem representations.

\subsubsection{Symmetry breaking in the classical model}
\label{sec:PrimalModel}
\begin{figure}[htb]
\begin{center}
\includegraphics[width=0.5\textwidth]{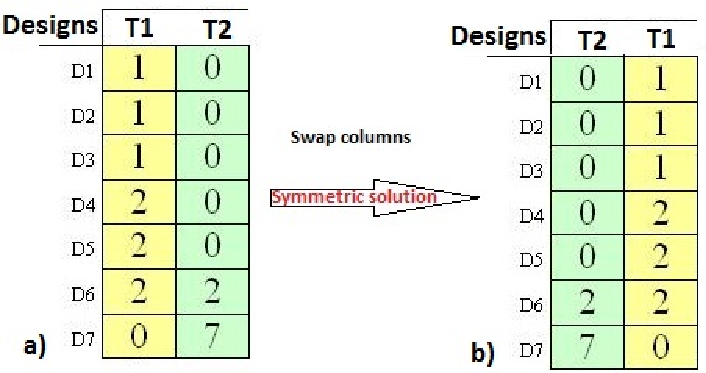}
\end{center}
\caption{Classical model: Example of a candidate encoding (a) and a symmetrical individual (b) for some problem instance $\text{TDP}\langle 7,2,9\rangle$}
\label{fig:ModA.SB}
\end{figure}

In the classical model ($P$), symmetries appear at the template level. So, given a candidate $c$, a symmetrical candidate can be easily obtained by simply interchanging two templates (i.e., columns) in the classical encoding of $c$. Figure \ref{fig:ModA.SB} shows, for a problem instance $\text{TDP}\langle 7,2,9\rangle$, an example of a candidate representation (left) and a symmetrical equivalent (right). Note that these are, in someway, equivalent  to the encoding shown in Figure \ref{fig:TDPModB}. For instance, there are two pairs of designs 4, 5 and 6, and one occurrence of designs 1, 2, 3 in the same template; in the other template one can find seven occurrences of variation 7 and two of variation 6. 

To remove this type of symmetry, we propose imposing a lexicographic order for the configurations of each template in a solution candidate. So, let $c$ be a candidate solution represented in the classical encoding $M=\{s_{ij}\}_{v\times t}$ for a given problem instance $\text{TDP}\langle v,t,s\rangle$. The configuration of a template 
$T_j$ (for a column $j \in \{1,t\}$) in $c$ is defined as the tuple (or sequence) of values that compounds column $j$ in $M$, that is to say:
$$\overline{config_c(j)} = \langle s_{1j},\ldots,s_{vj}\rangle$$
Then, a lexicographic order between templates is imposed as a new constraint in the problem $\varphi$ (see Section \ref{subsec:primalmodel}), that is to say:

\begin{equation}
 \forall j \in \{1,t-1\}:\ \overline{config_c(j)} \leq_{lex}  \overline{config_c(j+1)}
\label{primal model: symmetry breaking constraint}
\end{equation}

This lexicographic ordering constraint $\leq_{lex}$  is implemented as follows. Let $\overline{a} = \langle a_1,\ldots,a_v\rangle$ and $\overline{b} = \langle b_1,\ldots,b_v\rangle$ be two tuples of values representing two distinct template configurations. Then

\begin{displaymath}
 \overline{a} \leq_{lex} \overline{b} = \text{true} \text { if }  \left\{
     \begin{array}{c}
        a_1 < b_1, \text{ or} \\
        a_1=b_1 \text{ and } \langle a_2,\ldots,a_v\rangle \leq_{lex}  \langle b_2,\ldots,b_v\rangle
     \end{array}
     \right\} \text{false, otherwise }
\end{displaymath}

Note that this new constraint imposes a lexicographic order by columns that removes template symmetries. For instance, Figure \ref{fig:ModA.SB}(a) does not represent a valid candidate as,  for $j=1$, $\overline{config_c(1)} \not\leq_{lex}  \overline{config_c(2)}$. That is to say, $\langle 1,1,1,2,2,2,0 \rangle \not\leq_{lex} \langle 0,0,0,0,0,2,7 \rangle$ (i.e., column 1 is not lexicographically less thatn or equal to column 2). However, Figure \ref{fig:ModA.SB}(b) is a feasible candidate.

\subsubsection{Symmetry breaking in the alternative model}
\label{sec:BreakingSDualModel}

\begin{figure}[htb]
\begin{center}
\includegraphics[width=0.4\textwidth]{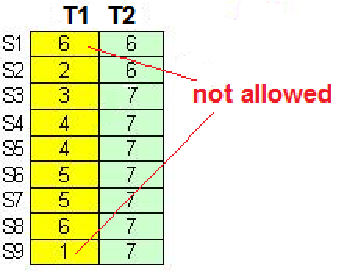}
\end{center}
\caption{Alternative model: Example of an individual that is initially symmetrical to that shown in Figure \ref{fig:TDPModB}. Imposing the constraint shown in Equation (\ref{dual:symmetry breaking constraint}) removes the possibility of symmetry.}
\label{fig:ModB.NoPermite}
\end{figure}

The alternative model has more symmetries than the classical model. We have a template symmetry  that is removed in the same way as in the classical model. Additionally the interchanging of two values (i.e., variations) in any column (i.e., any template)  produces symmetrical individuals. For instance, Figure  \ref{fig:ModB.NoPermite} shows a symmetrical representation of the candidate displayed in Figure \ref{fig:TDPModB}. The two only differ in the exchange of values stored in positions $M_D[1,1]$ and $M_D[1,9]$ (i.e., values $s_{11}$ and $s_{19}$). This symmetry can be broken by imposing an increasing ordering in each column, that is to say:

\begin{equation}
  \forall j \in \{1,t\}: (\forall i \in \{1,s-1\}: s_{ij} \leq s_{i+1,j})
 \label{dual:symmetry breaking constraint}
\end{equation}

So, the encoding shown in Figure  \ref{fig:ModB.NoPermite} is not valid
as it does not satisfy constraint (\ref{dual:symmetry breaking constraint}). However, the representation of that individual shown in Figure \ref{fig:TDPModB} is a feasible candidate as it satisfies the constraints shown in Equations (\ref{primal model: symmetry breaking constraint}) and (\ref{dual:symmetry breaking constraint}).

%% file: hybridalgorithms.tex
\section{Hybrid metaheuristics to deal with the problem}
\label{NewBattery}

This section describes a number of hybrid metaheuristics that we have implemented in this paper to deal with the TDP. These metaheuristics  are constructed from other metaheuristics. The idea is that these  hybrid methods allow the synergistic union of different techniques working on our two problem representation proposals (i.e., classical and alternative) and considering the possible employment of symmetry breaking.

\input{memeticAlgorithms}

\input{cooperativeAlgorithms}

%% file: memeticAlgorithms.tex
\subsection{Memetic algorithms}
\label{Sec:hybridization}

Memetic algorithms are a type of hybrid-optimisation methods which, in its classical version, consists in  executing a local search inside the execution of an external genetic algorithm (GA)\citep{cotta16memetic,Neri2012book}. The general scheme of this MA is depicted below in Algorithm~\ref{fig:MA}. The input of this algorithm is constituted by  the problem parameters (i.e., $\langle v,t,s \rangle$), as well as the  algorithm's own data, i.e., the genetic operators and their associated parameters (such as application rates). The output of the algorithm is the best individual found during the search process. Selection is done using binary tournament for breeding and replacement of the worst individual in the population. To keep diversity in the population, duplicated solutions are not accepted, and a re-starting mechanism is introduced to re-activate the search whenever it stagnates. Local search is restricted to exploring $n_{\nu}$ neighbours, which in other words means evaluating $n_{\nu}$ candidates. Local improvement strategy has been executed immediately after the  mutation stage (Line 15). Local search is applied individually according to a probability $\mathit{p_{LS}}$. When this local improvement is applied, the algorithm is executed over a number of $\mathit{Evals_{LS}}$ evaluations of the objective function. This is a very simple improvement strategy. Note that there are other, more complex techniques in the literature to determine what kind of improvement,  as well as to which individual of the population this improvement should be applied \citep{yew_keane,yew_Lim_Zhu}. 

\begin{algorithm}[h]
\scriptsize{
\caption{Pseudo code of the memetic algorithm.}
\label{fig:MA}
\Begin {
\For {$i \leftarrow 1$ \textbf{to} $popsize$}
{
	$pop[i]$ $\leftarrow$ \textsc{GenerateMatrix}($V,T,s$)\;
	\textsc{Evaluate}($pop[i]$)\;
}
\While {$numEvals < maxEvals$}
{
	\eIf {rand $<p_{X}$}
	{
		$parent_1$ $\leftarrow$ \textsc{TournamentSelect}($pop$)\;
		$parent_2$ $\leftarrow$ \textsc{TournamentSelect}($pop$)\;
		$\mathit{offspring}$ $\leftarrow$ \textsc{Recombine}($parent_1$, $parent_2$)\;
	}
	{
		$\mathit{offspring}$ $\leftarrow$ \textsc{TournamentSelect}($pop$)\;
	}
	$\mathit{offspring}$ $\leftarrow$ \textsc{Mutate}($\mathit{offspring}, p_M$)\;
	\lIf {rand $<p_{LS}$}
	{
		$\mathit{offspring}$ $\leftarrow$ \textsc{LocalSearch}($\mathit{offspring}, n_{\nu}$)
	}
	\textsc{Evaluate}($\mathit{offspring}$)\;
	$pop$ $\leftarrow$ \textsc{Replace}($pop$, $\mathit{offspring}$)\;
	\lIf {stagnation($n_\iota$)} {
		$pop$ $\leftarrow$ \textsc{Restart}($pop$, $f\%$)
	}
	
}
}}
\end{algorithm}

In this paper, we propose to employ MAs to handle the TDP. Note however that the concept of MA is not new
and has already been applied successfully to other manufacturing problems \cite{DBLP:journals/jim/LiKT03,DBLP:journals/jim/JolaiAASO11}.
However, to the best of our knowledge,  our proposal represents the first application of an MA to deal with the TDP.

%% file: cooperativeAlgorithms.tex
\subsection{Cooperative algorithms}
\label{Sec:cooperative algorithms}

Memetic algorithms, in the aforementioned form, can be considered integrative algorithms in the sense that a metaheuristic is integrated inside another metaheuristic. In this paper, we also consider another kind of hybridisation, as explained in the introduction section, in the form of collaborative methods in which several metaheuristics perform in an isolated way and interchange information from time to time (perhaps synchronously, but not necessarily so). In this paper, we propose to tackle the TDP through the parallel execution of a number of metaheuristics that are executed independently (most probably exploring different parts of the search space). These metaheuristics synchronise, from time to time, to interchange information (basically solutions found) between pairs of metaheuristics (also called agents in this paper, as this schema mimics a network of agents, which cooperate and where each agent represents a metaheuristic). This approach has  proven  to be efficient for a number of combinatorial problems \citep{Cruz2009,Masegosa2009,Amaya2011MC}.

In this paper we consider cooperative algorithms that connect  a preset number of those metaheuristics, previously described in this paper (i.e., local search, genetic algorithms or memetic algorithms, and their versions adapted to cope with the classical/alternative formulation and with the possible imposition of symmetry breaking). Note that this means that we can consider techniques working on different search spaces (e.g., classical and alternative) and even metaheuristics which, in addition, may (or not) impose symmetry breaking constraints. In other words,  some of the connected  metaheuristics in the cooperative algorithms  might possibly work on different encoding/search spaces and/or manage distinct policies to exploit the symmetrical nature of the problem. 

 In  \cite{Amaya2011MC} we handled another combinatorial optimisation problem with collaborative approaches and obtained acceptable results (i.e., some success). However, these approaches  did not take into account symmetry breaking nor different encodings, nor representation spaces. Our cooperative algorithms are designed from a similar point of view and with to the following schema of execution: the idea is to apply a number of (possibly different) optimisation algorithms each of which explores a  specific part of the search landscape through processes of intensification.
The agents synchronise from time to time to exchange information. In other words, the agents engage in periods of isolated exploration followed by synchronous communication in which some (or all) agents share solution candidates. The aim is to reactivate or improve the subsequent search process that each agent will carry out in the subsequent phase of isolated executions.  

More specifically, initially, each metaheuristic connected in the cooperative algorithm is initialised with random solutions 
Then, the cooperative algorithm is executed for a maximum number of synchronisation cycles $\Theta$, where in each cycle each metaheuristics is executed independently, keeping (and updating)  its own pool of solutions (i.e., the best solutions or candidates found so far by the specific metaheuristic). At the end of each cycle, the metaheuristics (also called agents) share information. This basically means that solutions are transmitted from one agent to another according to certain communication topology, which is identified by a specific spatial structure  (e.g. solutions are transmitted from any given agent to its successor in a ring-based structure).

 Three different communication topologies have been considered here (the same as in \cite{Amaya2011MC}): (\textsc{Ring}), in which there exists a circular list of metaheuristics in which each one only sends (resp.
	receives) information to its successor (resp. from its predecessor); (\textsc{Broadcast}), in which the best
	overall solution at each synchronisation point is transmitted to all the other techniques in collaboration; and (\textsc{Random}), in which the information (i.e., the best candidate solution found so far in a metaheuristic before the synchronisation step) is transmitted from one metaheuristic to another, which has been previously randomly chosen.

In addition, we have adapted some of the ideas proposed in \cite{NoguerasCotta2014} for memetic algorithms to our cooperative algorithms and have considered a number of  policies for the submission and acceptance of  candidates. This basically means that the metaheuristic that submits information to another metaheuristic, has to choose the candidate to transmit (from its solution pool) according to the following three policies: (\textsc{Random} R), that is to say,  send a random solution from its solution pool; (\textsc{Diverse} D), that is to say, send the candidate that maximises the diversity\footnote{\label{pie1}To this end, individuals, whose genotypic distance (in a Hamming sense) to individuals in the receiving population is maximal, are selected.} , and (\textsc{Worst} W), in which the worst candidate from its solution  pool is submitted.

In addition, the metaheuristic which receives a solution candidate, from another method in the synchronisation steps, has to decide whether to accept it or not. In the case it accepts it, the incoming solution has to replace one of the candidates in the method's own solution poo. These decisions are taken on the basis of the reception and replacement policies associated with the metaheuristic.  
As for the reception and replacement policies in the destination metaheuristics, three alternatives are also considered: (\textsc{Random} R): always accept the submitted candidate and replace one random individual in its pool; (\textsc{Diverse} D): accept a new individual if and only if, it improves the diversity of its solution pool and replace the worst, and (\textsc{Worst} W): always accept the candidate and replace the worst in pool. 

Note that the cooperative algorithm has several parameters associated with the problem (i.e., $V$, $T$, specific demands for each variation, etc.), the topology of the agent network (which defines the communication policy, as explained below),  the number $n$ of metaheuristics connected, the candidate migration policy, the  criteria for accepting the candidates, and the number of communication cycles $\Theta$. Each metaheuristic  also has its own parameters (such as, for instance, operator application rates, and type of encoding i.e., classical or alternative).

At the end of the last cycle of execution, the cooperative algorithm returns the best individual found by any of the metaheuristics, linked collaborativelly. 

Note that these cooperatives schemas are not novel but they are, for the  first  time,  used here  to tackle the TDP. In addition, the particularity of this paper, is that each agent/metaheuristic can work on the classical or alternative formulation, with or without breaking symmetry constraints. Moreover, the metaheuristics can also be integrative techniques (i.e., an MA) combining an LS and a GA also working on different search spaces and with distinct policies for symmetry breaking.

%% file: experiments.tex
\section{Experiments}
\label{sec:experiments}
This section describes the experimental analysis conducted to check the validity of our proposals\footnote{In order to favour transparent research, and encourage comparison with other research, both the data employed in the work presented here  and the source code of the algorithms generated from it have been deposited in a public repository: \url{https://github.com/drrueda/TDP} (August 2019).}. We have considered a large number of algorithms, which have been generated by combining all the proposals described in the preceding sections of the paper. We first describe in detail the notation used to denote the algorithms involved in the experiments, as well as the experimental setting in Sections \ref{sec:notation} and \ref{sec:configuration}, respectively. The results obtained in an experimental evaluation of all the techniques are shown in Section~\ref{subsect:Algoexperiments} and, subsequently, a comparison of their performance, on a statistical level, is made in Section~\ref{subsect:experimental comparison and result analysis}.

\input{notation}

\input{configuration}

\input{experiments_basic_hybrid}

\input{experimental_comparison}

%% file: notation.tex
\subsection{Notation}
\label{sec:notation}
Each specific algorithm is associated with a sequence of identifiers, separated by dots, that allow the nature of the technique in question to be easily identifiable. To begin with, the basic metaheuristics  are hill climbing (Hc), tabu search (Ts) and the genetic algorithm (GA). For the experiments, we have employed the versions that we described in \citep{RodriguezCottaFernandezMAEB2010}, with the same parameter configurations.
Additionally, for the population-based methods -- i.e., the GA and the memetic algorithm (MA) described in Section \ref{Sec:hybridization} --  we have used the same versions (with the same parameter configurations, as well) that we described in \citep{RodriguezCottaFernandezIMACS2011}.Specifically, we consider the same two recombination procedures, that is to say, uniform crossover (Ux) and greedy crossover (Gd). For GA and MA, we have also considered here multi-parent recombination and the notation A$m$ is used to denote its arity, i.e., the number $m$ of parents used in the reproduction step. Additionally, we use an asterisk ($*$) to indicate the use of symmetry-breaking methods (the absence of an asterisk indicates no symmetry-breaking techniques were employed), and P (resp. D) to indicate that the search was conducted on the primary or classical (resp. alternative) representation space.

{\em Examples of notations.} Below, we provide some examples of notations of the algorithms considered in the experiments: for instance, 

\begin{itemize}
\item Hc.P (resp. Hc.P*) denotes a hill climbing method that was implemented for the classical model without (resp. with) symmetry breaking,

\item  Ts.D (resp. Ts.D*)  denotes a tabu search implemented for the alternative model without (resp. with) symmetry breaking (as explained in Section \ref{sec:BreakingSDualModel}),

\item likewise, Ga.D*.A2.Ux denotes a genetic algorithm, with 2-parent uniform crossover, implemented for the alternative encoding with symmetry breaking,

\item In addition, Ga.D.A4.Gd is a genetic algorithm with a 4-parent greedy crossover implemented for the alternative formulation without symmetry breaking, 

\item Ma.Hc.P.A2.Gd denotes a memetic algorithm with a hill climbing method as local search  and a 2-parent greedy  recombination implemented for the classical formulation without symmetry breaking, and

\item MA.Ts.D*.A2.Ux is a memetic algorithm with tabu search as the improvement method and a 2-parent uniform crossover operator implemented for the alternative model with symmetry breaking.
\end{itemize}

For more details on the LS techniques and GA, and the greedy and uniform crossover operators, the reader is referred to  \citep{RodriguezCottaFernandezMAEB2010,RodriguezCottaFernandezIMACS2011}.

Regarding cooperative methods, these algorithms are composed of some of the previous techniques combined according to certain topology and migration policies. The notation $\mathbf{T}n(a_1,\ldots,a_n)$MR is used to characterise the method. Here:

\begin{itemize}
\item $\mathbf{T} \in \{\textsc{Broadcast}$ (Bc), $\textsc{Random}$ (Ra), $\textsc{Ring}$ (Ri)$\}$ denotes the topology of the model, 
\item ${n}$ is the number of agents (i.e. metaheuristics) connected cooperatively, 
\item ${a}_i$ is the metaheuristics used by agent $i$ (for $1 \leq i \leq n$), and
\item M, R $\in \{\textsc{Random}$ (R), $\textsc{Diverse}$ (D), $\textsc{Worst}$ (W) $\}$ identify,  respectively, the policies to migrate and accept candidates in the agents (see Section~\ref{Sec:cooperative algorithms}). In our experiments, we have considered the following six combinations for $migration-reception$ policies:
\textsc{Random}-\textsc{Random} (RR), 
\textsc{Random}-\textsc{Worst} (RW),
\textsc{Random}-\textsc{Diverse} (RD),
\textsc{Diverse}-\textsc{Random} (DR),
\textsc{Diverse}-\textsc{Worst} (DW), and 
\textsc{Diverse}-\textsc{Diverse} (DD).
Note that we do not include the combinations WD (i.e. \textsc{Worst}-\textsc{Diverse}), WR (i.e. \textsc{Worst}-\textsc{Random}) and WW (i.e. \textsc{Worst}-\textsc{Worst}). The reason is that preliminary experiments showed that choosing the \textsc{Worst} policy for migration exhibited a poor performance compared to the other combinations. 
\end{itemize}

{\em Examples of notation of cooperative algorithms}: Ra2(Ts.P, MA.Ts.D.A2.Gd)RW is a 2-agent \{\textsc{random} topology\}-based cooperative algorithm that connects (a) a TS working on the classical representation, and (b) an MA that works on the alternative representation, which uses a 2-parent greedy crossover, and that integrates TS as the underlying local search; in this case,  the algorithm always sends a random candidate selected from  the solution pool of the origin node (i.e., a \textsc{Random} policy for migration), which will replace the worst individual in the destination node (i.e., a \textsc{Worst} policy for the acceptance policy). Similarly, Ri3(Ts.P, MA.Ts.P.A2.Gd, MA.Ts.D.A4.Gd)RD denotes a 3-agent cooperative algorithm that connects, in a \textsc{ring} topology, (a) tabu search and (b) two different MAs; the individuals to migrate are randomly chosen (i.e. a \textsc{Random} policy for migration) whereas candidates are accepted only if they increase the diversity of the solution pool (i.e. \textsc{Diverse} acceptance criteria). 

Note that in the cooperative algorithms the same optimisation method may be used by several agents (this is the case, for instance, in the algorithm Bc4(Ts.P,Ts.P,Ts.P,MA.Ts.D*.A4.Gd)RD in which 3 of the 4 agents contain the local search Ts.P.). The rationale for this is to try to increase the contribution of a certain method to the resulting cooperative hybrid, whose overall search profile is influenced by the particular mix of optimisation methods used. 
For clarity, in these cases,
we use the notation $\mathbf{T}n(pa, qb)$MR to denote the $n$-agent cooperative algorithm 
\[\mathbf{T}n(\underbrace{a,\ldots,a}_{p\ {\rm times}}, \underbrace{b \ldots b}_{q\ {\rm times}}){\rm MR}\] 
in which agents $a$ and $b$ are employed $p$ and $q$ times respectively (and where $p$ and $q$ are arbitrary numbers  that fulfil $n=p+q$);  moreover, $p$ (resp. $q$) is not written when $p=1$ (resp. $q=1$). So, for instance, Bc4(3Ts.P, MA.Ts.D*.A4.Gd)RD denotes the model Bc4(Ts.P,Ts.P,Ts.P,MA.Ts.D*.A4.Gd)RD  (i.e. here $p=3$ and $q=1$). Also,
Ra5(3Ts.P,2MA.Ts.P.A2.Gd)DW is a 5-agent algorithm where the local search Ts.P is embedded in 3 agents and the algorithm MA.Ts.P.A2.Gd is contained within the other two agents (i.e. here $p=3$ and $q=2$). 

The entire set of algorithms considered in the experiments should be easily identifiable from the notation introduced in this section.

%% file: configuration.tex
\subsection{Experimental configuration} 
\label{sec:configuration}
The experiments were conducted on three problem instances taken from \cite{Proll97ilpand}, see Table \ref{tab:Escenarios}. 
All algorithms were run 20 times per problem instance. However (as shown later) the performance of the cooperative metaheuristics was so promising that we decided to  increase the number of runs per problem instance in order to assess the robustness of these algorithms. So, given the number $n$ of metaheuristics involved in a cooperative algorithm, this was run $n \times 10$ times per problem instance.

For all the metaheuristics, the number of evaluations for each scenario was dependent on the number of variations and templates: $n_{\nu}=1000 \cdot t \cdot v\cdot(v-1)\cdot {\%}_v$, where ${\%}_v$ represents the percentage of neighbors to be evaluated. Note that a full exploration schema of the neighbourhood is very costly. For instance,  for $v=50$ and $t = 4$, evaluating just $5\%$ of the neighborhood means evaluating  $4.9 \cdot 10^5$ neighbors, therefore a partial exploration policy was considered. That being said, all the methods consider the equivalent number of full evaluations in each case. This means, that all the metaheuristics involved in our experimental evaluation consume exactly the same number of evaluations (as indicated previously for each problem instance) during their executions. This ensures a fair comparison of performance. In this sense, and considering fairness, note that, in the cooperative algorithms, given a global number of evaluations $E_{max}$ for a specific problem instance --as indicated above-- and a specific number $\Theta$ of interaction cycles, the number of evaluations that can be consumed in each cycle is $E_{cycle} = E_{max}/\Theta$ evaluations. In addition, note also that for a given number $n$ of  metaheuristics connected in the algorithm, this means that each metaheuristic consumes $E_{cycle}/n$ evaluations per cycle.

The number of evaluations without improvement to trigger intensification in a local search method or re-starting in a population-based method is $n_\iota = n_\nu/10$. Other parameters of the population-based algorithms (i.e., GA or MA) are \emph{population size}$=100$, crossover and mutation probabilities $p_X=.9$ and $p_M=1/\ell$ (where $\ell=v\cdot t$ is the size of individuals in the classical model and  $\ell=s\cdot t$ is the size of individuals for the alternative model) respectively, and $f_\%=10\%$ (i.e., $l_i=u_i=0.10$ for any variation $i$). In the case of the MA, $p_{LS}$ is set to 0.005.

\begin{table}[htb]
\caption{Problem instances (taken from \cite{Proll97ilpand}).}
\label{tab:Escenarios}
%\begin{small}
\begin{center}
\scalebox{0.85} 
{
\begin{tabular}{lccl}
Problem & Slots per Template & Variations & Demand (x1000) \\
\hline
(A) Cat Food Cartons & 9 & 7 & 250,255,260,500,500,800,1100 \\
\hline
(B) Herbs Cartons & 42 & 30 &  60,60,70,70,70,70,70,70,70,80, \\
&    &    &  80,80,80,90,90,90,90,90,90,100, \\
&    &    &  100,100,100,150,230,230,230,230,280,280 \\
\hline
(C) Magazine Inserts & 40 & 50 & 50,53,55,60,85,90,100,100,105,110, \\
&    &    & 137,140,140,140,150,150,150,150,150,150,\\
&    &    & 150,150,168,170,170,195,195,200,200,200,\\
&    &    & 210,210,225,230,230,230,250,250,250,250,\\
&    &    & 250,250,250,250,265,270,270,375,375,405\\
\hline
\end{tabular}}
\end{center}
%\end{small}
\end{table}

Specifically for the cooperative algorithms, the idea is to harness the synergy between the metaheuristics when these work in cooperation. We have considered the three topologies proposed (ring, broadcast and random) with a number $n$ of agents between 2 and 5, and a number of cycles for the synchronisation (i.e., transmission of information between agents/metaheuristics)  $5$ (we set this value based on preliminary experiments  with values un $\{5,10,15 \}$).
These parameter values were chosen because some preliminary experiments indicated that they provided a good trade-off between the computational cost and the quality of solutions attained. 

As for the  versions of these algorithms executed in the alternative representation of the problem, all of them used the same parameters (population size, genetic operator rates, \ldots, etc) as their equivalent classical versions. In addition, versions with symmetry breaking follow the requirements described in  Sections~\ref{sec:PrimalModel} and~\ref{sec:BreakingSDualModel} for the classical and alternative representations, respectively. The combination of the two problem-representation models and the possibility of breaking the symmetries give way to four different  scenarios, namely,  classical representation with/without symmetry breaking, and its equivalents in the alternative model with/without symmetry breaking.

%% file: experiments_basic_hybrid.tex
\subsection{Experimental results}
\label{subsect:Algoexperiments}

In this paper we have considered a high number of metaheuristics that have been applied to cope with the TDP. In the following we list them, and show their performance in dealing with the TDP.

\subsubsection{Basic approaches}

Twenty-four algorithms have been considered in the experimentation, i.e., 8 local searches (resulting from the two LS methods considered in this paper -- HC and TS-- combined with the four aforementioned scenarios), and 16 genetic algorithms
(resulting from implementing  GA in each of the four  scenarios referred to above, combined with the use of  two distinct crossover operators (i.e., Ux/Gd), and the recombination of 2 or 4 parents). More specifically, the 24 basic metaheuristics that we have coded are:
Hc.P,
Hc.P*,
Hc.D,
Hc.D*,  
Ts.P,
Ts.P*,
Ts.D,
Ts.D*,
Ga.P.A2.Gd,  
Ga.P*.A2.Gd,  
Ga.P.A2.Ux,   
Ga.P*.A2.Ux, 
Ga.D.A2.Gd,  
Ga.D*.A2.Gd,
Ga.D.A2.Ux ,
Ga.D*.A2.Ux,  
Ga.P.A4.Gd, 
Ga.P*.A4.Gd,  
Ga.P.A4.Ux,  
Ga.P*.A4.Ux,  
Ga.D.A4.Gd,  
Ga.D*.A4.Gd,  
Ga.D.A4.Ux,
and  
Ga.D*.A4.Ux . 

\begin{table}[!htb]
\caption{Basic approaches: Number (and percentage rate) of  feasible solutions found by the  top 5 ranked basic algorithms and nondominated basic versions (from the set of 24 basic methods considered) working on the problem instances taken from \cite{Proll97ilpand}. Rows are ordered according to the rank value assigned to each algorithm (shown in fifth column). The far-right column identifies the algorithms that are not dominated  by any other techniques (i.e., those marked with the symbol \XSolidBrush) according to the tuple $\langle n_{catfood},n_{herbs},n_{magazine}\rangle$, where $n_{catfood}$, $n_{herbs}$ and $n_{magazine}$ corresponding to the number of solutions found in the first, second and third problem instances, respectively. In general, a  solution $s$ is nondominated if there is no other solution that improves any of the objective values of $s$ and is no worse in the remaining objective values. }
\label{tab:NumSolIndAll}
\begin{center}
\scalebox{0.85}
{
\begin{tabular}{|c|c|c|c|c|c|}
\hline
Metaheuristics & Cat Food & Herbs C. & Magazine I. & Average ranking & N.D \\ 
\hline
Hc.D*  &    \bf{20}  (      100.00  \% )    &     1  (        5.00  \% )    &     0  (        0.00  \% ) &  4.33 & \XSolidBrush\\
Hc.P*  &    \bf{20}  (      100.00  \% )    &     1  (        5.00  \% )    &     0  (        0.00  \% ) &   4.67& \XSolidBrush\\
Ga.P.A4.Gd  &    17  (       85.00  \% )    &    \bf{18}  (       90.00  \% )    & 0  (    0.00  \% )    &  6.00& \XSolidBrush\\
Ga.D*.A4.Gd  &    {19}  (       95.00  \% )    &     2  (       10.00  \% )   &    0 (  0.00  \% )    &  6.33& \XSolidBrush\\ 
Ts.P  &    19  (       95.00  \% )    &     1  (        5.00  \% )    &     0  (        0.00  \% )       &  6.50 &\\
Ga.P*.A4.Gd  &    16  (       80.00  \% )    &     0  (        0.00  \% )    &     1  (   5.00  \% )     &  7.00& \XSolidBrush\\ 
Ga.P*.A2.Gd  &    14  (       70.00  \% )    &     0  (        0.00  \% )    &   \bf{3}(  15.00  \% )    &  8.33& \XSolidBrush\\ 
\hline
\end{tabular}}
\end{center}
\end{table}

Due to the high number of algorithm variants it is not easy to analyse the performance of each of them compared with the rest by simply inspecting the numerical tables, so we have opted for a rank-based approach. More precisely, we have computed the rank $r_j^i$ of each algorithm $j$ on each instance $i$. For the purpose of ranking them, in all cases we have used the sum of the number of feasible solutions found on each problem instance (from  the set of 20 runs). The best algorithm receives rank 1 and the worst one receives rank $k$, where $k=24$ is the number of algorithms involved in the ranking. 
To simplify the amount of data to show, Table \ref{tab:NumSolIndAll} gives the performance results obtained by the  top five ranked techniques. This table also includes  
the basic algorithms that are not dominated  by any other basic metaheuristic. The concept of dominance (as used in this paper) is based on the {\em Pareto dominance} presented in multiobjective optimisation \citep{DBLP:reference/nc/Zitzler12}. More precisely in the context of this paper, maximising the number of solutions found for each problem instance might be considered as an objective inside a multiobjective approach.The fifth column of Table~\ref{tab:NumSolIndAll} indicates the average ranking value obtained from the distributions of the rankings of each algorithm. For each of the 7 basic algorithms (identified in the first column) is also shown the number of times that a problem instance was solved in its 20 runs (in columns two to four -- in brackets -- the corresponding success percentage is also given). 

In general the TS, HC and GA variants perform very well in the instance of lowest complexity. Also, at first sight, the  algorithms {\bf{\it{Hc.D*}}} perform better than their counterparts, local search (i.e., TS). 
More specifically, the GA algorithm working on the classical model with a 4-parent greedy crossover and using the greedy operator (i.e., Ga.P.A4.Gd) has been shown to be highly efficient for the problem instances with lowest and medium complexity. 
In general terms, this GA  finds the highest number of solutions, exactly 35 (from 60 runs, 20 runs per problem instance), which represents a success percentage of close to 60\%.
However, the most complex scenario remains unsolvable for this algorithm. In general, the {\em Magazine Insert} instance is very hard for all the basic methods although it was solved by some GA variants, those that manage symmetry constraints, but, which curiously could not solve the {\em Herbs Cartoon} problem instance. In fact, the management of symmetry constraints in the classical formulation  is a common feature of the four algorithms that did solve this instance.

\subsubsection{Integrative methods}

 Thirty-two hybrid integrative algorithms (i.e., memetic algorithms) have been considered in the experimentation. This number results from considering two distinct representations (i.e., classical/alternative), two  recombination operators (i.e., UX/Gd), two local improvement techniques (i.e., HC and TS), the possibility of applying  symmetry-breaking procedures,  and the selection of 2 or 4 parents for recombination. All the possible combinations produce 32 distinct scenarios (i.e., $2 \times 2 \times 2 \times 2 \times 2$) that are associated with 32 different memetic algorithms. 

More specifically, these 32 MAs are the following:
Ma.Hc.P.A2.Gd,
Ma.Hc.P*.A2.Gd,  
Ma.Hc.D.A2.Gd,
Ma.Hc.D*.A2.Gd,  
Ma.Hc.P.A2.Ux,  
Ma.Hc.P*.A2.Ux,  
Ma.Hc.D.A2.Ux, 
Ma.Hc.D*.A2.Ux, 
Ma.Hc.P.A4.Gd,  
Ma.Hc.P*.A4.Gd,  
Ma.Hc.D.A4.Gd,  
Ma.Hc.D*.A4.Gd,  
Ma.Hc.P.A4.Ux,  
Ma.Hc.P*.A4.Ux,  
Ma.Hc.D.A4.Ux,  
Ma.Hc.D*.A4.Ux,  
Ma.Ts.P.A2.Gd,  
Ma.Ts.P*.A2.Gd,
Ma.Ts.D.A2.Gd, 
Ma.Ts.D*.A2.Gd, 
Ma.Ts.P.A2.Ux, 
Ma.Ts.P*.A2.Ux,  
Ma.Ts.D.A2.Ux,  
Ma.Ts.D*.A2.Ux,  
Ma.Ts.P.A4.Gd,
Ma.Ts.P*.A4.Gd,     
Ma.Ts.D.A4.Gd,  
Ma.Ts.D*.A4.Gd,
Ma.Ts.P.A4.Ux,  
Ma.Ts.P*.A4.Ux,  
Ma.Ts.D.A4.Ux , and 
Ma.Ts.D*.A4.Ux 

Again, as done previously to simplify the amount of data to show, Table~\ref{tab:NumSolforMAsAll} just shows the performance results obtained by the  top five ranked techniques, and also   includes the memetic versions that are not dominated  by any other memetic metaheuristics.  In general the memetic methods perform reasonably well in all the problem instances. In fact, more than $59\%$ of these algorithms find feasible solutions in all of them. Considering the sum of the number of feasible solutions found,  the best algorithms are Ma.Hc.P*.A2.Gd, Ma.Hc.P*.A2.Ux, Ma.Hc.P*.A4.Gd and Ma.Ts.P.A2.Ux.

\begin{table}[htb]
\caption{Integrative hybrid algorithms: Number (and percentage rate) of  feasible solutions found by the  selected memetic algorithms (from a set of 32 memetic versions) working on the problem instances taken from \cite{Proll97ilpand}. Rows are ordered according to the rank value assigned to each algorithm (shown in the fifth column).
As in Table \ref{tab:NumSolIndAll}, the far-right column identifies the algorithms that are not dominated  by any other techniques (i.e., those marked with the symbol \XSolidBrush) according to the tuple $\langle n_{catfood},n_{herbs},n_{magazine}\rangle$, where $n_{catfood}$, $n_{herbs}$ and $n_{magazine}$ correspond to the number of solutions found in the first, second and third problem instance respectively.}
\label{tab:NumSolforMAsAll}
\begin{center}
\scalebox{0.85}
{
\begin{tabular}{|c|c|c|c|c|c|}
\hline
Metaheuristics & Cat Food & Herbs C. & Magazine I. & Average ranking & N.D \\ 
\hline
Ma.Hc.P*.A2.Ux  &    17  (       85.00  \% )    &     \bf{19  (   95.00  \% )}    &     \bf{10  ( 50.00  \% )}   &     4.83& \XSolidBrush\\ 
Ma.Hc.P*.A4.Gd  &    17  (       85.00  \% )    &    18  (       90.00  \% )    &     8  (       40.00  \% )     &     6.00&\\ 
Ma.Ts.P*.A2.Gd  &    17  (       85.00  \% )    &     \bf{19  (       95.00  \% )}    &     6  (   30.00  \% )   &     6.67&\\ 
Ma.Hc.P*.A2.Gd  &    17  (       85.00  \% )    &    17  (       85.00  \% )    &     9  (       45.00  \% )     &     7.17&\\ 
Ma.Ts.P.A2.Ux  &    17  (       85.00  \% )    &     7  (       35.00  \% )    &     9  (       45.00  \% )      &     8.00&\\ 
Ma.Hc.D.A4.Gd  &     \bf{18  (       90.00}  \% )    &     0  (        0.00  \% )    &     7  (    35.00  \% )   &     13.67& \XSolidBrush\\ 
Ma.Ts.P.A2.Gd  &     \bf{18  (       90.00  \% )}    &     1  (        5.00  \% )    &     0  (     0.00  \% )   &     17.17& \XSolidBrush\\ 
\hline
\end{tabular}}
\end{center}
\end{table}		

At first glance, and considering the number of solutions found, MAs clearly outperform the basic approaches that have been applied to the problem. The best MA (i.e., Ma.Hc.P*.A2.Ux), according to the number of solutions found and the rank-based classification shown in the fith column in Table~\ref{tab:NumSolforMAsAll}, obtained 46  solutions (in 60 runs) which represents a success rate of more than 80\%.  In addition, the hardest problem instance (i.e., Magazine I.) is solved (at least, in one run) by 29 MAs (out of 32 MA versions). Moreover, the best MA  was succesful in exactly half of its runs. Note that those MAs, which manage symmetry constraints and are executed in the classical model perform reasonably well in the hardest instance.

In the basic and memetic approaches, we have identified those techniques that are not dominated by any other method from its category (i.e., basic or memetic) because they are later used  in a global comparison. The far-right columns in Tables \ref{tab:NumSolIndAll} and \ref{tab:NumSolforMAsAll} mark the nondominated algorithms for the classes of basic and integrative methods, respectively.

\subsubsection{Cooperative techniques}

Note the huge amount of collaborative methods that can be considered from the template $\mathbf{T}n(a_1,\ldots,a_n)$MR as explained previously (see Section~\ref{sec:notation}) as any $a_i$ can be any of the basic or memetic metaheuristics described in this paper which gives rise to a huge number of combinations. 
To have a more manageable set of algorithms, we decided to focus our experimets on cooperative algorithms that have exactly two metaheuristics (although these  can be duplicated in the method).  In other words, we have considered cooperative algorithms that are compounded by $p$ instances of a metaheuristic $a$ and $q$ instances of a metaheuristic $b$. This basically means that our cooperative methods are devised from the aforementioned template  $\mathbf{T}n(p a,q b)$MR (see Section~\ref{sec:notation}) where $p+q=n$ and $a,b$ are metaheuristics belonging to a given collection ${\cal A}_i$ that contains two types of agents.  In this paper, we have considered  the following four collections:

\begin{itemize}
\item ${\cal A}_1 =$ \{Ts.D,MA.Hc.P*.A2.Ux\}
\item ${\cal A}_2 =$\{Ts.D,GA.D*.A4.Gd\}
\item ${\cal A}_3 =$\{Ts.D,MA.Ts.P.A2.Gd\} 
\item ${\cal A}_4 =$\{MA.Hc.P*.A2.Ux,MA.Ts.P.A2.Gd\}
\end{itemize}

The algorithms in these collections have been picked due to their good individual performances according to Tables~\ref{tab:NumSolIndAll} and~\ref{tab:NumSolforMAsAll}, or because we try to impose diversity (in the sense of fostering the collaboration of metaheuristics of distinct kinds) in the cooperation. So, each collection represents a different way to combine methods. For instance  ${\cal A}_1$ contains two methods  with different characteristics because one basic, focused on the alternative encoding, and with no symmetry breaking, whereas the other is a population-based method, searching in the classical encoding search space and with symmetry breaking (i.e., ${\cal A}_1$ represents a model D-P*). Collection ${\cal A}_2$ represents the cooperation among techniques simultaneously working in the alternative formulation (i.e., a model D-D*), ${\cal A}_3$  the cooperation among methods working on distinct computation domains without symmetry breaking (i.e.,a  model P-D), and ${\cal A}_4$ the methods that work in the same computation domain (in this case, the classical one) but with different policies for symmetry breaking  (i.e., a model  P-P*).

Considering all possible combinations of the three topologies, a number of agents between two and five (i.e., four possibilities), six distinct  combinations of migration/reception policies (see Sections~\ref{Sec:cooperative algorithms} and~\ref{sec:notation}), and four collections of agents, a total of 288 algorithmic variants were created (i.e., $3 \times 4 \times 6 \times 4$). 

From this wide set of methods, a relatively high number of methods showed an improvement with respect to the other metaheuristics that we had previously considered. Moreover, 17 cooperative algorithms exhibited a success rate (i.e., percentage of successful runs with respect to the total number of runs) of above 70\% with one close to 95\%. Table \ref{tab:InsResuTop17TDP} shows these methods,
and Table \ref{tab:Number of  Solutions of the coperative methods above 70 percent of success} displays the number of  solutions found for each of these collaborative methods in the $n \times 10$ runs of the algorithms in each problem instance.

%-----------------------------------------------------------------------------
\begin{table}[!h]
	\caption{(Central column) percentage (\%) who managed to find more than 70\% of feasible solutions, in scenarios from Table \ref{tab:Escenarios}  solved by those cooperative algorithms (identifed in first column). The right-hand column shows the collection of algorithms that collaborate in the cooperative search.}
	\label{tab:InsResuTop17TDP}
	\begin{center}
%		\scalebox{0.8}{
			\begin{tabular}{|c|c|c|}
				\hline
				Algorithms & $>$ \% Fact. & Collection\\
				\hline
				Ra2(Ts.D,Ga.D*.A4.Gd)DW & 70.00 \% & ${\cal A}_2$\\ 
				Bc5(Ts.D,Ga.D*.A4.Gd)RD & 72.00 \% & ${\cal A}_2$\\ 
				Bc4(Ts.D,Ma.Ts.P.A2.Gd)RD & 72.50 \% & ${\cal A}_3$\\ 
				Ri3(Ts.D,Ma.Hc.P*.A2.Ux)RD & 73.33 \% & ${\cal A}_1$\\ 
				Ri5(Ts.D,Ga.D*.A4.Gd)RD & 74.00 \% & ${\cal A}_2$\\ 
				Ra5(Ts.D,Ma.Ts.P.A2.Gd)RD & 74.00 \% & ${\cal A}_3$\\ 
				Ra5(Ts.D,Ga.D*.A4.Gd)RD & 76.00 \% & ${\cal A}_2$\\ 
				Ra3(Ts.D,Ma.Hc.P*.A2.Ux)RD & 76.67 \% & ${\cal A}_1$\\ 
				Bc4(Ts.D,Ma.Hc.P*.A2.Ux)RD & 77.50 \% & ${\cal A}_1$\\ 
				Bc4(Ts.D,Ga.D*.A4.Gd)RD & 77.50 \% & ${\cal A}_2$\\ 
				Ri5(Ts.D,Ma.Ts.P.A2.Gd)RD & 78.00 \% & ${\cal A}_3$\\ 
				Bc2(Ts.D,Ma.Hc.P*.A2.Ux)RD & 80.00 \% & ${\cal A}_1$\\ 
				Ri5(Ts.D,Ma.Hc.P*.A2.Ux)RD & 80.00 \% & ${\cal A}_1$\\ 
				Bc5(Ts.D,Ma.Ts.P.A2.Gd)RD & 82.00 \% & ${\cal A}_3$\\ 
				Ri3(Ts.D,Ga.D*.A4.Gd)RD & 83.33 \% & ${\cal A}_2$\\ 
				Bc2(Ts.D,Ma.Ts.P.A2.Gd)RD & 85.00 \% & ${\cal A}_3$\\ 
				Bc5(Ts.D,Ma.Hc.P*.A2.Ux)RD & 94.00 \% & ${\cal A}_1$\\
				\hline 
		\end{tabular}
%}
	\end{center}
\end{table}

\begin{table}[!h]
\caption{Cooperative algorithms: number and percentage of solutions found in each of the three problem instances by the cooperative methods that reached, at least, 70\% of success in the 60 runs (20 runs per problem  instance)}
\label{tab:Number of  Solutions of the coperative methods above 70 percent of success}
\begin{center}
%\scalebox{0.7}{
\begin{tabular}{|c|c|c|c|}
\hline
algorithm 			& \# (\%) (Cat Food) & \# (\%) (Herbs)  & \# (\%) (Magaz.) \\ 	
\hline
Bc2(Ts.D,Ma.Hc.P*.A2.Ux)RD & 20.00 (100.00 \%) & 18.00 (90.00 \%) & 16.00 (80.00 \%) \\ 
Ra3(Ts.D,Ma.Hc.P*.A2.Ux)RD & 30.00 (100.00 \%) & 23.00 (76.67 \%) & 23.00 (76.67 \%) \\ 
Ri3(Ts.D,Ma.Hc.P*.A2.Ux)RD & 30.00 (100.00 \%) & 24.00 (80.00 \%) & 22.00 (73.33 \%) \\ 
Bc4(Ts.D,Ma.Hc.P*.A2.Ux)RD & 40.00 (100.00 \%) & 35.00 (87.50 \%) & 31.00 (77.50 \%) \\ 
Bc5(Ts.D,Ma.Hc.P*.A2.Ux)RD & 50.00 (100.00 \%) & 48.00 (96.00 \%) & 47.00 (94.00 \%) \\ 
Ri5(Ts.D,Ma.Hc.P*.A2.Ux)RD & 50.00 (100.00 \%) & 42.00 (84.00 \%) & 40.00 (80.00 \%) \\ 
Ra2(Ts.D,Ga.D*.A4.Gd)DW & 20.00 (100.00 \%) & 14.00 (70.00 \%) & 14.00 (70.00 \%) \\ 
Ri3(Ts.D,Ga.D*.A4.Gd)RD & 30.00 (100.00 \%) & 25.00 (83.33 \%) & 25.00 (83.33 \%) \\ 
Bc4(Ts.D,Ga.D*.A4.Gd)RD & 40.00 (100.00 \%) & 34.00 (85.00 \%) & 31.00 (77.50 \%) \\ 
Bc5(Ts.D,Ga.D*.A4.Gd)RD & 50.00 (100.00 \%) & 40.00 (80.00 \%) & 36.00 (72.00 \%) \\ 
Ra5(Ts.D,Ga.D*.A4.Gd)RD & 50.00 (100.00 \%) & 48.00 (96.00 \%) & 38.00 (76.00 \%) \\ 
Ri5(Ts.D,Ga.D*.A4.Gd)RD & 50.00 (100.00 \%) & 39.00 (78.00 \%) & 37.00 (74.00 \%) \\ 
Bc2(Ts.D,Ma.Ts.P.A2.Gd)RD & 20.00 (100.00 \%) & 18.00 (90.00 \%) & 17.00 (85.00 \%) \\ 
Bc4(Ts.D,Ma.Ts.P.A2.Gd)RD & 40.00 (100.00 \%) & 34.00 (85.00 \%) & 29.00 (72.50 \%) \\ 
Bc5(Ts.D,Ma.Ts.P.A2.Gd)RD & 50.00 (100.00 \%) & 46.00 (92.00 \%) & 41.00 (82.00 \%) \\ 
Ra5(Ts.D,Ma.Ts.P.A2.Gd)RD & 50.00 (100.00 \%) & 45.00 (90.00 \%) & 37.00 (74.00 \%) \\ 
Ri5(Ts.D,Ma.Ts.P.A2.Gd)RD & 50.00 (100.00 \%) & 45.00 (90.00 \%) & 39.00 (78.00 \%) \\ 
\hline
\end{tabular}
%}
\end{center}
\end{table}

Table\ref{tab:NumPorcxCriterioTop17TDP} gives an idea of the influence of each of the design parameters of these 17 cooperative algorithms(considered the best cooperative methods). Note that strategy RD has a high impact, and topology  BROADCAST is used in more than half of the methods. In addition, the number of 
agents/metaheuristics that seem to offer the most robust performance is five.

\begin{table}[!t]
	\caption{Relative frequency of each particular design parameter among the 17 cooperative algorithms from Table \ref{tab:InsResuTop17TDP}. Left-hand column indicates the combination MR of migration(M)/reception(R) policies where  M, R $\in \{\textsc{Random}$ (R), $\textsc{Diverse}$ (D), $\textsc{Worst}$ (W) as explained in Sect. \ref{Sec:cooperative algorithms}. Central column shows the communication topology where Bc = \textsc{Broadcast}, Ra = \textsc{Random}, and Ri = \textsc{Ring}. Right-hand column refers the number of agents in the algorithm. }
	\label{tab:NumPorcxCriterioTop17TDP}
	\begin{center}
%		\scalebox{0.7}{
			\begin{tabular}{crrlcrrlcrr}
				\hline
				\multicolumn{3}{c}{M/R Policy} & ~\hspace{2mm}~ & \multicolumn{3}{c}{Topology} & ~\hspace{2mm}~ & \multicolumn{3}{c}{Number of agents} \\
				\cline{1-3}\cline{5-7}\cline{9-11}
				DW 	& ~1 &(5.88 \%) 	&& Bc	& 8 &(47.06 \%) && $n=2$	& ~3 &(17.65 \%) \\
				RD	& 16 &(94.12 \%) 	&& Ra	& 4 &(23.53 \%) && $n=3$	& ~3 &(17.65 \%) \\
				&    &	             && Ri	& 5 &(29.41 \%) && $n=4$	& ~3 &(17.65 \%) \\
				&    &	&&		&	&		                && $n=5$	& ~8 &(47.05 \%)\\
				\hline
		\end{tabular}
%}
	\end{center}
\end{table}

As mentioned we have also compared the algorithms in a rank-based approach and have applied tests of  Friedman and Iman-Davenport, finding the existence of significant differences among these 17 cooperative algorithms at the standard level  (i.e., $\alpha = 0.05$).
We have also carried out a Holm-Bonferroni test to determine whether there are significant differences with respect to a control algorithm,
in this case, Bc5(Ts.D,Ma.Hc.P*.A2.Ux)RD (i.e., the algorithm with the best average rank according to the rank-based classification). The results are shown in Table~\ref{tab:holmCooperaTop17TDP}. Note that there are significant statistical differences  with respect to six algorithms but not with respect to ten of them.

\begin{table}[!ht]
	\caption{Results of the Holm-Bonferroni test, for cooperative algorithms using \textsf{Bc5(Ts.D,Ma.Hc.P*.A2.Ux)RD}  as control algorithm, at the standard level of $\alpha = 0.05$. Only the algorithms that show no significant differences with respect to the control algorithm are shown (i.e. those for which p-value $\ge \alpha/i$).}
	\label{tab:holmCooperaTop17TDP}
	\begin{center}
%		\scalebox{0.7}{
			\begin{tabular}{lcccc}
				\hline
				$i$ & algorithm & z-statistic  &  p-value &       $ \alpha / i $ \\ 
				\hline
				1 & \textsf{Ra5(Ts.D,Ga.D*.A4.Gd)RD} & 5.255e-01 & 2.996e-01 & 5.000e-02\\ 
				2 & \textsf{Bc5(Ts.D,Ma.Ts.P.A2.Gd)RD} & 8.085e-01 & 2.094e-01 & 2.500e-02\\ 
				3 & \textsf{Ri5(Ts.D,Ma.Hc.P*.A2.Ux)RD} & 8.085e-01 & 2.094e-01 & 1.667e-02\\ 
				4 & \textsf{Ri5(Ts.D,Ma.Ts.P.A2.Gd)RD} & 1.011e+00 & 1.561e-01 & 1.250e-02\\ 
				5 & \textsf{Ra5(Ts.D,Ma.Ts.P.A2.Gd)RD} & 1.091e+00 & 1.375e-01 & 1.000e-02\\ 
				6 & \textsf{Ri5(Ts.D,Ga.D*.A4.Gd)RD} & 1.172e+00 & 1.205e-01 & 8.333e-03\\ 
				7 & \textsf{Bc5(Ts.D,Ga.D*.A4.Gd)RD} & 1.374e+00 & 8.466e-02 & 7.143e-03\\ 
				8 & \textsf{Bc4(Ts.D,Ga.D*.A4.Gd)RD} & 2.021e+00 & 2.163e-02 & 6.250e-03\\ 
				9 & \textsf{Bc4(Ts.D,Ma.Hc.P*.A2.Ux)RD} & 2.142e+00 & 1.608e-02 & 5.556e-03\\ 
				10 & \textsf{Bc4(Ts.D,Ma.Ts.P.A2.Gd)RD} & 2.385e+00 & 8.541e-03 & 5.000e-03\\ 
				\hline	
		\end{tabular}
%}
	\end{center}
\end{table}

%% file: experimental_comparison.tex
\subsection{Experimental comparison and result analysis}
\label{subsect:experimental comparison and result analysis}

Finally, we perform a cross-comparison between all the methods considered so far. Due to the high number of methods to compare, we have decided that the selection of basic and integrative methods to be included in the comparison should be  based on the concept of {\em Pareto dominance} as already indicated, that is to say, those methods marked 
in the far right columns in Tables \ref{tab:NumSolIndAll} and \ref{tab:NumSolforMAsAll} for  the classes of basic and integrative methods, respectively. Specifically, we have selected six basic methods  from Table \ref{tab:NumSolIndAll} and three techniques from Table \ref{tab:NumSolforMAsAll}. With respect to the collaborative techniques, we have included the set of 11 methods that did not show statistical differences, as shown in Table \ref{tab:holmCooperaTop17TDP} (i.e., the ten algorithms displayed in the table plus the control algorithm).

\begin{figure}[!h]
	\label{torneo:20NoPasanHolm.Integra+CooperaTDP}
	\begin{center}
		\includegraphics[scale=.6]{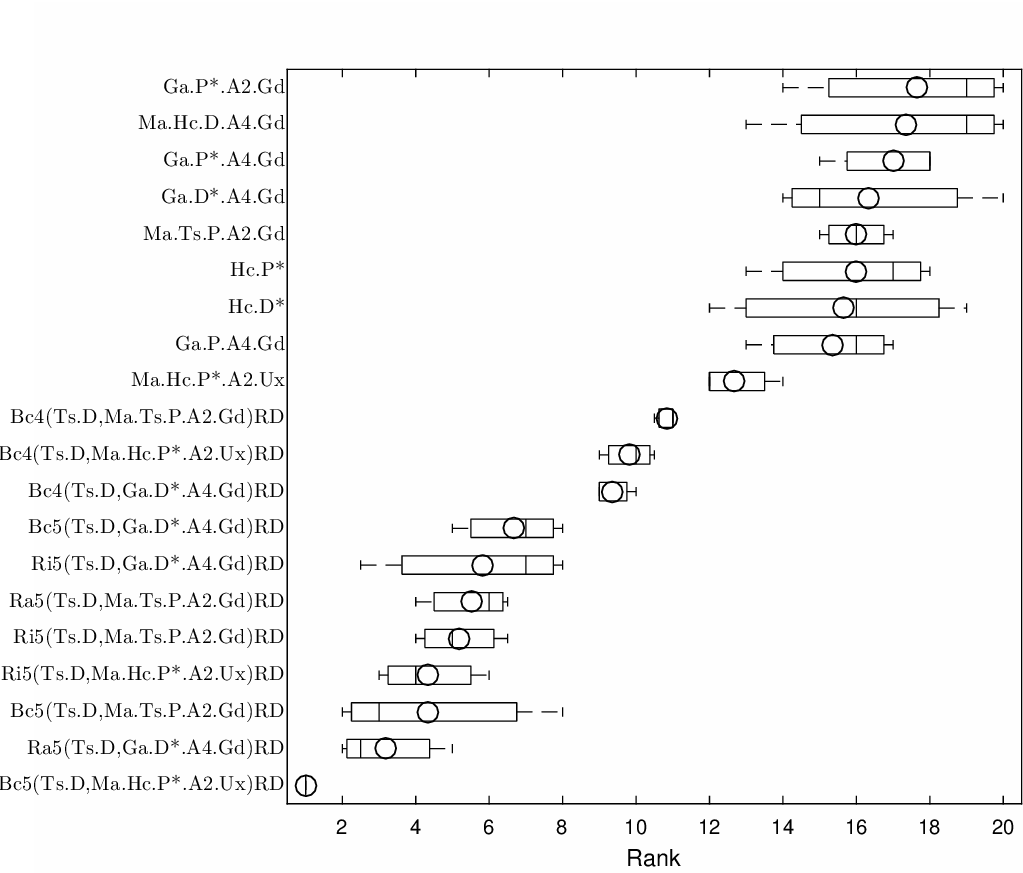}
	\end{center}
	\caption{Rank distribution for the integrative and cooperative techniques chosen because of their performance in the experimentation}.
\end{figure}

We have once again used  a rank-based approach to compare the performance of these 20 algorithms, and the results are shown in Figure~\ref{torneo:20NoPasanHolm.Integra+CooperaTDP}. Observe that Bc5(Ts.D,Ma.Hc.P*.A2.Ux)RD (i.e., the algorithm with five metaheuristics from collection ${\cal A}_1$,  executing in parallel and synchronysing  via a brodcast topology, and random-diverse as the  candidate migration-reception strategy) is the  best ranked algorithm. Note that all cooperative algorithms are ranked in the best positions and that the algorithmic versions with symmetry breaking
tend to rank better. This is an important result that supports the usefulness of symmetry breaking to improve solving capabilities in the TDP, which could lead to the design of other algorithms to deal with this problem.

\begin{table}[!h]
	\caption{Results of the Friedman and Iman-Davenport tests ($\alpha=0.05$) for the integrative and cooperatives techniques better positioned in the experimentation.}
	\label{test:friedman20NoPasanHolm.Integra+CooperaTDP}
	\begin{center}
%		\scalebox{0.7}{
			\begin{tabular}{ccccc}
				\hline
				& ~~~ Friedman ~~~ & ~~~ Critical $\chi^2$ ~~~ & ~~~Iman-Davenport~~~ & ~~~Critical \\
				& value                    &   value      &        value          & $F_F$ value \\
				\hline  
				All 20	 & 51.557143	 & 30.143527	 & 18.944882	 & 1.867332 \\ 
				\hline
		\end{tabular}
%}
	\end{center}
\end{table}

The application of the tests of  Friedman and Iman-Davenport show the existence of significant statistical differences among the techniques compared. We can observe that the critical values are lower than the values obtained in the respective tests, -- see Table \ref{test:friedman20NoPasanHolm.Integra+CooperaTDP} --.  We have conducted a Holm-Bonferroni's test with  {\textit{Bc5(Ts.D,Ma.Hc.P*.A2.Ux)RD}} as control algorithm; see Table \ref{test:Holm20NoPasanHolm.Integra+CooperaTDP}.
The results highlight that there are no significant differences with the cooperative methods nor with the memetic algorithm \textsf{Ma.Hc.P*.A2.Ux}, but there are significant differences with respect to the other eight metaheuristics (i.e., GAs and MAs).

\begin{table}[!h]
	\caption{Results of the Holm test ($\alpha=0.05$) for the integrative and the cooperative techniques better positioned in the experimentation, using \textsf{Bc5(Ts.D,Ma.Hc.P*.A2.Ux)RD} as control algorithm. Only the algorithms that showed no significant differences with respect to the control algorithm are shown (i.e. those for which p-value $\ge \alpha/i$)}
	\label{test:Holm20NoPasanHolm.Integra+CooperaTDP}
	\begin{center}
%		\scalebox{0.8}{
			\begin{tabular}{lcccc}
				\hline
				$i$ & algorithm & z-statistic  &  p-value &       $ \alpha / i $ \\ 
				\hline
				1 & \textsf{Ra5(Ts.D,Ga.D*.A4.Gd)RD} & 4.485e-01 & 3.269e-01 & 5.000e-02\\ 
				2 & \textsf{Bc5(Ts.D,Ma.Ts.P.A2.Gd)RD} & 6.901e-01 & 2.451e-01 & 2.500e-02\\ 
				3 & \textsf{Ri5(Ts.D,Ma.Hc.P*.A2.Ux)RD} & 6.901e-01 & 2.451e-01 & 1.667e-02\\ 
				4 & \textsf{Ri5(Ts.D,Ma.Ts.P.A2.Gd)RD} & 8.626e-01 & 1.942e-01 & 1.250e-02\\ 
				5 & \textsf{Ra5(Ts.D,Ma.Ts.P.A2.Gd)RD} & 9.316e-01 & 1.758e-01 & 1.000e-02\\ 
				6 & \textsf{Ri5(Ts.D,Ga.D*.A4.Gd)RD} & 1.001e+00 & 1.585e-01 & 8.333e-03\\ 
				7 & \textsf{Bc5(Ts.D,Ga.D*.A4.Gd)RD} & 1.173e+00 & 1.204e-01 & 7.143e-03\\ 
				8 & \textsf{Bc4(Ts.D,Ga.D*.A4.Gd)RD} & 1.725e+00 & 4.225e-02 & 6.250e-03\\ 
				9 & \textsf{Bc4(Ts.D,Ma.Hc.P*.A2.Ux)RD} & 1.829e+00 & 3.372e-02 & 5.556e-03\\ 
				10 & \textsf{Bc4(Ts.D,Ma.Ts.P.A2.Gd)RD} & 2.036e+00 & 2.089e-02 & 5.000e-03\\ 
				11 & \textsf{Ma.Hc.P*.A2.Ux} & 2.415e+00 & 7.863e-03 & 4.545e-03\\ 
				\hline
		\end{tabular}
%}
	\end{center}
\end{table}

Subsequently, we have carried out a one-to-one comparison between the best ranked algorithm {\textit{Bc5(Ts.D,Ma.Hc.P*.A2.Ux)RD}}, according
 to  Figure~\ref{torneo:20NoPasanHolm.Integra+CooperaTDP}, and each of the 11 techniques for which no significant statistical differences are found according to Table \ref{test:Holm20NoPasanHolm.Integra+CooperaTDP}.  The results are shown in Table \ref{tab:Ranksum:12:no:PasanHolmItegra+Coopera}. Here, we can observe that there are significant differences with the best memetic algorithm Ma.Hc.P*.A2.Ux. Note also that there are, at least, 6 cooperative techniques that do not show significant differences with respect to the best cooperative technique. This result highlights the superiority of the cooperative metaheuristics in tackling the TDP.

\begin{table}[!h]
	\caption{Head to head comparison of \textsf{Bc5(Ts.D,Ma.Hc.P*.A2.Ux)RD} with the remaining algorithms on each of the three problem	instances. Each entry in the table contains three symbols corresponding (from left to right) to Cat Food Cartons, Herbs Cartons and Magazine Inserts: $\bullet$ (resp. $\circ$) indicates that the difference in performance on the corresponding instance is (resp. is not) statistically significant at $\alpha=0.05$ using a ranksum test.}
	\label{tab:Ranksum:12:no:PasanHolmItegra+Coopera}
	\begin{center}
		\scalebox{0.9}{
			\begin{tabular}{ccc}
				\hline
				Ma.Hc.P*.A2.Ux  &  Ra5(Ts.D,Ga.D*.A4.Gd)RD  &  Bc5(Ts.D,Ma.Ts.P.A2.Gd)RD \\ 
				%\hline 
				$\bullet$/$\bullet$/$\bullet$ & $\bullet$/$\circ$/$\circ$ & $\bullet$/$\circ$/$\circ$ \\ 
				\hline
				 Ri5(Ts.D,Ma.Hc.P*.A2.Ux)RD  & Ri5(Ts.D,Ma.Ts.P.A2.Gd)RD  &  Ra5(Ts.D,Ma.Ts.P.A2.Gd)RD \\
				$\bullet$/$\bullet$/$\circ$ & $\bullet$/$\circ$/$\circ$ & $\bullet$/$\circ$/$\circ$  \\ 
				%\hline
				\hline  
				 Ri5(Ts.D,Ga.D*.A4.Gd)RD & Bc5(Ts.D,Ga.D*.A4.Gd)RD & Bc4(Ts.D,Ga.D*.A4.Gd)RD \\ 
				%\hline
				$\bullet$/$\bullet$/$\circ$ & $\bullet$/$\bullet$/$\bullet$ & $\bullet$/$\bullet$/$\bullet$ \\ 
				%\hline 
				\hline 
				 Bc4(Ts.D,Ma.Hc.P*.A2.Ux)RD  &  Bc4(Ts.D,Ma.Ts.P.A2.Gd)RD  &   \\ 
				%\hline
				 $\bullet$/$\circ$/$\circ$ & $\bullet$/$\circ$/$\circ$ & \\ 
				\hline
		\end{tabular}
}
	\end{center}
\end{table}

In general, all the hybrid metaheuristics (including the integrative and cooperative approaches) show an acceptable performance to tackle the TDP, having representatives that can solve all the problem instances, even the most complex ones. However, the cooperative metaheuristics considered in this paper to tackle the TDP, are more efficient (in terms of both success rate and speed in finding solutions - this means that the cooperative methods consume fewer evaluations to find a solution than the rest of the metaheuristics) than their constituent parts (i.e., the basic and memetic algorithms) working alone in the solving of the most complex problem instances. Moreover, cooperative methods also improved the quality of the solutions. For instance, for  Herbs Cartoon, the best solution found by a cooperative technique  saves  $548$ units in the final product with respect to the  best solution found by the memetic algorithms. Moreover, for the instance  Magazine, the saving is of $31.500$ units. See Tables ~\ref{tab:majorConfig3scenarios} (best solution founds by integrative techniques) and~\ref{tab:majorConfig3scenariosTDP} (best solutions found by cooperative techniques) 

%\clearpage
\begin{table}[htb]
%\begin{adjustwidth}{-2.25in}{0in}
\caption{Best solutions found by Ma.Hc.P*.A2.Ux in the 3 problem instances (minimal waste using the less number of templates). }
\label{tab:majorConfig3scenarios}
\begin{center}
%\begin{sideways}
	\scalebox{0.6}{
\begin{tabular}{|cccccccccc|}
\hline
 Problem &  & No. Template & Templates &~& Pressings &~& Overall Desv. & T.Desv & Waste\\ 
\hline
Cat Food  Cartons & &   2  & [1,1,1,2,2,2,0] & &     250000 &&  & &\\
&  &  &  [0,0,0,0,0,2,7] & &     157143 &&         0.80 &        -3.85/        1.79 &        29287 \\
\hline
Herbs Cartons  & &   2  & [1,1,1,1,1,1,1,1,1,1,1,1,1,1,1, & & && & & \\
&   & &  0,1,1,1,1,1,1,1,2,2,2,3,3,4,4] & &      65911 &&  & &\\
&  &  &  [0,0,0,0,0,0,0,0,0,1,1,1,1,1,1, & & && & & \\
&   & &  6,1,1,1,2,2,2,2,1,6,6,2,2,1,1] & &      16363 &&         2.99 &        -8.58/        9.85 &       104548 \\
\hline
Magazine Inserts  & &   3  & [1,1,1,1,1,1,2,2,2,2,2,0,0,0,0,0,0,0,0,0,0,0,0,0,0, & & && & & \\
&   & &  1,1,1,1,1,1,1,1,1,1,1,2,2,2,2,2,2,0,0,0,0,1,0,0,0] & &      54000 &&  & &\\
&  &  &  [0,0,0,0,1,1,0,0,0,0,1,0,0,0,0,0,0,0,0,0,0,0,1,1,1, & & && & & \\
&   & &  0,0,0,0,0,0,0,1,1,1,1,0,0,0,0,0,0,3,3,3,3,2,2,2,12] & &      33750 &&  & &\\
&   & &  [0,0,0,0,0,0,0,0,0,0,0,1,1,1,1,1,1,1,1,1,1,1,1,1,1, & & && & & \\
&   & &  1,1,1,1,1,1,1,1,1,1,1,1,1,1,1,1,1,1,1,1,1,1,2,2,0] & &     146000 &&         2.97 &       -10.00/        8.00 &       277500 \\
\hline
\end{tabular}}
%\end{sideways}
\end{center}
%\end{adjustwidth}
\end{table}

\begin{table}[htb]
	%\begin{adjustwidth}{-2.25in}{0in}
	\caption{Cooperative techniques: the best solutions for the 3 problem instances (minimal waste using a fewest number of templates). }
	\label{tab:majorConfig3scenariosTDP}
	\begin{center}
		\scalebox{0.6}
		{
			\begin{tabular}{|cccccccccc|}
				\hline
				Problem &  & No. Template & Templates         &~~& Pressings &~~& Overall Desv. & T.Desv & Waste\\ 
				\hline
				Cat Food  & &   2            &   [0,0,0,0,0,2,7] & &     157143 &&             &                           &\\
				& &                &  [1,1,1,2,2,2,0] & &     250000 &&         0.80 &        -3.85/        1.79 &        29287 \\
				\hline
				Herbs Cartoon  & &   2  & [1,1,1,1,1,1,1,1,1,1,0,1,1,1,1, & & && & & \\
				&   & &  1,1,1,0,1,1,1,1,2,3,3,3,2,4,4] & &      66000 &&  & &\\
				&  &  &  [0,0,0,0,0,0,0,0,0,1,5,1,1,1,1, & & && & & \\
				&   & &  1,1,1,6,2,2,2,2,1,2,2,2,6,1,1] & &      16000 &&         2.97 &        -8.89/       10.00 &       104000 \\
				\hline
				Mz.Inserts  & &   3  & [0,0,0,0,0,0,0,0,0,0,0,1,1,1,1,1,1,1,1,1,1,1,1,1,1, & & && & & \\
				&   & &  1,1,1,1,1,1,1,1,1,1,1,1,1,1,1,1,1,1,1,1,1,1,2,2,0] & &     150000 &&  & &\\
				& &  &  [1,1,1,0,1,1,2,2,2,2,2,0,0,0,0,0,0,0,0,0,0,0,0,0,0, & & && & & \\
				&   & &  1,1,1,1,1,1,1,1,1,1,1,2,2,2,2,2,2,1,0,0,0,1,0,0,0] & &      50000 &&  & &\\
				&   & &  [0,0,0,2,1,1,0,0,0,0,1,0,0,0,0,0,0,0,0,0,0,0,1,1,1, & & && & & \\
				&   & &  0,0,0,0,0,0,0,1,1,1,1,0,0,0,0,0,0,1,3,3,3,2,2,2,12] & &      33000 &&         2.63 &        -9.09/       10.00 &       246000 \\
				\hline
		\end{tabular}}
	\end{center}
	%\end{adjustwidth}
\end{table}

%% file: conclusions.tex
\section{Conclusions and future work}
\label{sec:conclusions}
This paper has dealt with the TDP, a very hard combinatorial problem that has previously been approached through integer  linear programming and constraint programming. Recently,  we tackled this problem by means of metaheuristics. More  specifically, we employed local searches and genetic algorithms that showed a moderate success in the handling of the smallest instances of the problem but no significant performance in the solving of the most complex scenarios. However, metaheuristics are considered efficient methods that can find enough-quality solutions at a reasonable computational cost. For this reason, this paper has explored and analysed other metaheuristic approaches to deal with this template design problem. The motivation has been to test whether metaheuristics are suitable for tackling the problem. We have also tried to find solutions of high quality, even in the most complex scenarios, so as to encourage comparison with other techniques in the future.

In our research, we have considered a number of issues related to solution encoding, problem formulation, the symmetrical nature of the problem, and distinct approaches that foster collaboration among different heuristics. First, we have defined an alternative problem formulation with a slot-based representation 
 (which is an alternative to other slot-based representations and to the classical model with a variation-based encoding of the candidates). 
All our heuristics proposed in previous work were adapted to this alternative formulation and have been evaluated here experimentally. In addition, based on the highly symmetrical nature of the problem, we have considered the possibility of imposing standard symmetry breaking procedures --used in constraint and integer programming-- to reduce the search space of the problem (in both classical and alternative representations). Therefore, we have designed a number of algorithms that hybridise distinct metaheuristics with the aim of improving the performance of each of the linked metaheuristics working alone.We emphasise that our hybrid algorithms  are based on well-known concepts such as symmetry breaking and memetic algorithms. More specifically, we have employed memetic algorithms to design integrative metaheuristics, and the concept of multi-agent systems to produce cooperative versions of our metaheuristics.  

An experimental evaluation (and comparison) of all the metaheuristics proposed in this work, has shown that the hybrid metaheuristics exhibit a high performance to handle the TDP.
So, although the degree of benefit in using these procedures is highly dependent on other design decisions, a memetic algorithm (i.e., an integrative metaheuristic) using symmetry breaking on the alternative representation is a very acceptable option for tackling the TDP. 
We have also observed that integrative metaheuristic have  behaved outstandingly in solving scenarios of low and medium complexity, and faily effectively in more complex scenarios. 
However, the cooperative metaheuristics have outperformed the memetic approaches, at the same time that they have shown a high robustness. Moreover, one cooperative version has obtained a success rate close to 95\% in solving all the problem scenarios, including even the most complex problem instance. In fact, some of the cooperative metaheuristics presented here can be considered the state-of-the-art techniques  in solving the three problem instances reported in the literature. 
The robustness and high performance of the cooperative metaheuristics in handling the TDP  opens up a line of future work which aims to study other approaches for the synergetic combination of metaheuristics in the solving of this problem.

As for future work, we plan to test the performance of other metaheuristics to tackle the TDP. Another line of future research is to look for alternative representations/formulations  to the template design problem. We have already mentioned the possibility of considering asymmetric representative formulations (ARFs) as alternatives to the natural symmetric formulation of the problem. In the same line of work, redundant modelling, a well-known technique employed in constraint programming that combines alternative problem representations linked by channelling
constraints, might be an interesting approach to analyse \cite{DBLP:journals/constraints/ChengCLW99}.

In addition, note that  symmetry breaking (SB)  favours the performance of our metaheuristics, on both hybrid methods and basic techniques (including  local search and genetic algorithms). 
 However, some authors, such as S. Prestwich \cite{DBLP:journals/anor/Prestwich03}, have reported that the addition of symmetry breaking constraints to a model can produce a negative effect of  local search performance,  whereas other approaches have found that SB improves local search performance \cite{DBLP:conf/cp/Yokoo97}. S. Prestwich investigated this issue and found  that complex local search can reduce the negative effects of applying symmetry breaking \cite{10.1007/11493853_21}. Our metaheuristics (including local search) are based on specific neighbourhoods and robust recombination operators that were specifically tailored to the problem. Thus, our local search methods  might be considered, in some sense, complex, which could  justify that SB improves its performance. In any case, there are few approaches that connect metaheuristics and the symmetry breaking that we have considered in this paper (i.e., those  used primarily in constraint  and integer programming) and, as a consequence, the employment of this kind of SB in metaheuristics is poorly understood.  
Anyway, as mentioned in  \cite{DBLP:journals/tec/Prugel-Bennett04}, thinking in terms of symmetry breaking   allows some of the 
common notions about evolutionary algorithms (such as exploration versus exploitation, or diversity) to be redefined, and  therefore, this may help provide new methods
for improving their performance on hard optimisation problems.
This is an interesting line of future work that is beyond the scope of this paper.